\pdfoutput=1

\documentclass[11pt]{article}
\usepackage[]{acl}
\usepackage{times}
\usepackage{latexsym}
\usepackage[T1]{fontenc}
\usepackage[utf8]{inputenc}
\usepackage{microtype}
\usepackage{inconsolata}
\usepackage{graphicx}
\usepackage{multirow}
\usepackage{xspace}
\usepackage{booktabs}
\usepackage{amsmath}
\usepackage{subcaption}
\usepackage{enumitem}
\usepackage{bm}

\definecolor{customgray}{HTML}{eeefee} 
\usepackage{url}

\title{Specializing Large Language Models to Simulate\\ Survey Response Distributions for Global Populations}

\newcommand{\tubingen}{$^1$}
\newcommand{\cuhksz}{$^2$}
\newcommand{\ku}{$^3$}
\newcommand{\boc}{$^4$}

\author{Yong Cao\tubingen, Haijiang Liu\cuhksz, Arnav Arora\ku, Isabelle Augenstein\ku, \\ \textbf{Paul Röttger}\boc, \textbf{Daniel Hershcovich}\ku \\
{\tubingen}University of Tübingen, Tübingen AI Center \\
{\cuhksz}Wuhan University of Science and Technology  \\
{\ku}University of Copenhagen, {\boc}Bocconi University
 \\
\texttt{ yong.cao@uni-tuebingen.de}, 
\texttt{ dh@di.ku.dk}
}

\begin{document}
\maketitle
\begin{abstract}

Large-scale surveys are essential tools for informing social science research and policy, but running surveys is costly and time-intensive. If one could accurately simulate group-level survey results, this could be very valuable to social science research. Prior work has explored the use of large language models (LLMs) for simulating human behaviors, mostly through prompting. In this paper, we are the first to specialize LLMs for the task of simulating survey response distributions. As a testbed for this task, we use country-level results from two global cultural surveys. We devise a fine-tuning method based on first-token probabilities to minimize divergence between predicted and actual response distributions for a given question. Then, we show that this method substantially outperforms other methods and zero-shot classifiers, even on unseen questions, countries, and a completely unseen survey. While even our best models struggle with the task, especially on unseen questions, our results demonstrate the benefits of specialization for simulation, which may accelerate progress towards sufficiently accurate simulation in the future.
\end{abstract}

\begin{figure}[t]
    \centering
    \includegraphics[width=\columnwidth]{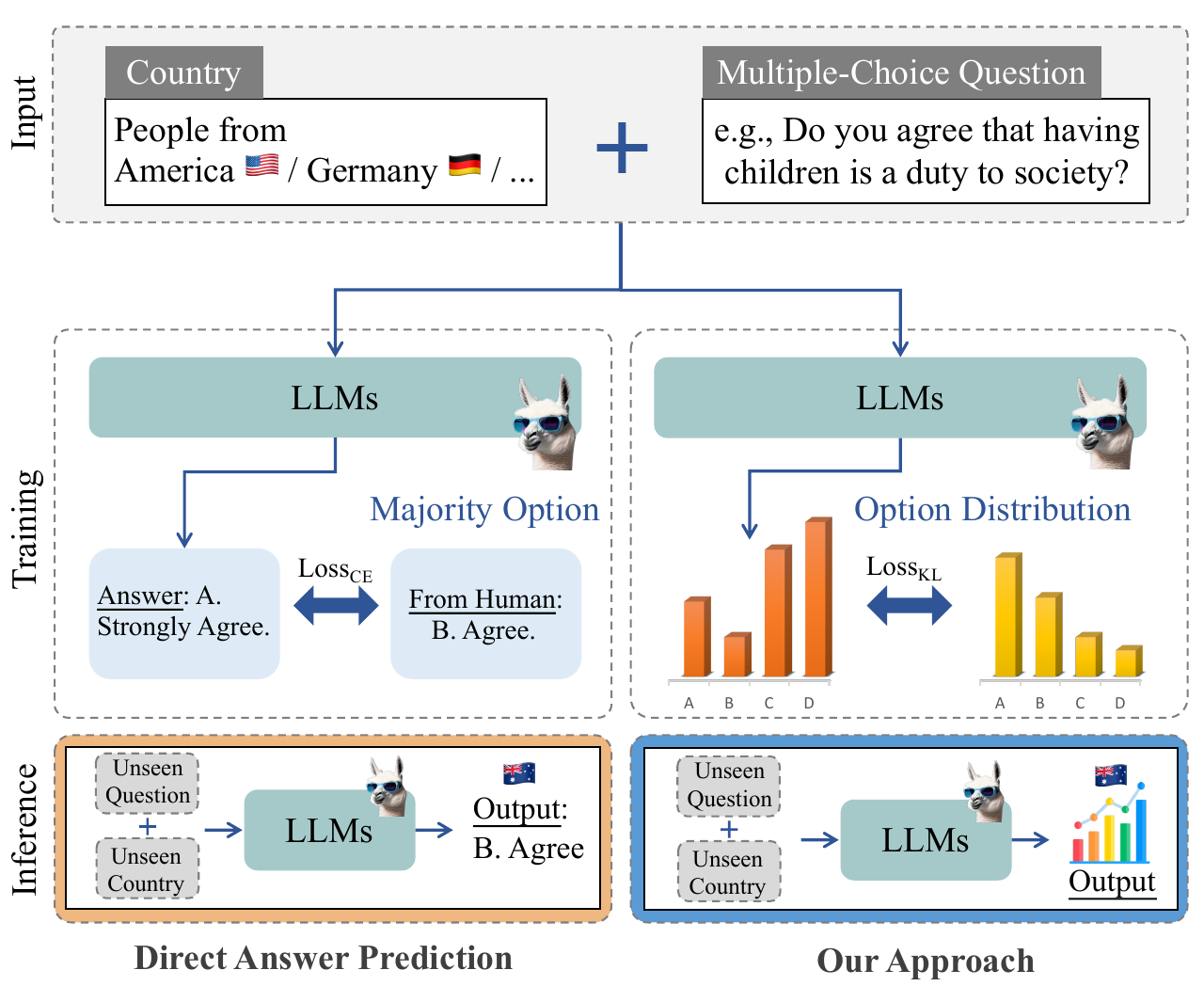} 
    \caption{\textbf{Overview of our proposed survey response distribution simulation framework} (right panel) versus direct answer prediction (left panel), highlighting a novel perspective on cultural simulation with LLMs.}
    \label{fig:figure_1}
\end{figure}

\section{Introduction}\label{sec:intro}
Humans are diverse, and they hold diverse opinions.
This is why surveys are essential tools for informing decision-making in policy and industry as well as social science research. 
Running large-scale surveys, however, is often costly and time-intensive.

Large language models (LLMs) have demonstrated promising potential for simulating human behaviors across groups and individuals \citep[][\textit{inter alia}]{argyle2023out,aher2023using, manning2024automated}.
LLM simulations of survey responses, if accurate towards the corresponding populations, could accelerate social science research and aid in more informed policy decisions.
Out of the box, however, LLMs are known to generate erroneous, stereotypical, or overconfident answers, especially in culturally diverse contexts \cite{yang2024calibrationmultilingualquestionanswering}, which limits their usefulness for survey simulations.
Prior work has at most tried to improve simulation accuracy through prompting strategies \citep{kwok2024evaluating,manning2024automated,sun2024random}.

In this study, our goal is instead to \textbf{\textit{specialize} LLMs for survey simulation} and gain a better understanding of how good LLMs can be at simulating survey responses when trained to do so, rather than how good they are when prompted out of the box.
As our main testbed, we use country-level response distributions from the widely-used World Values Survey \cite[WVS;][]{haerpfer2022world}.
When prompted with a survey question (e.g., ``In your opinion, should the use of nuclear power in Japan be reduced, maintained at its current level, or increased?''), corresponding answering options (e.g.\ ``Reduced'', ``Maintained at current level'', ``Increased'') and a target country (e.g.\ ``Japan''), we want our model to predict the distribution over the answering options for that target country.
To specialize LLMs for this task, we devise a fine-tuning method based on first-token probabilities, where the goal is to minimize divergence between predicted and actual country-level response distributions for a given survey question.

As shown in Figure~\ref{fig:figure_1}, we train models on one set of questions and countries from the WVS, then evaluate on both seen and unseen countries and questions as well as another completely unseen survey.
Across seven LLMs from three model families, our fine-tuning method substantially boosts prediction accuracy on seen and unseen WVS countries and questions.
These results also hold for a completely unseen survey, i.e. the Pew Global Attitudes Survey.
Simultaneously, we find the performance of even the best-fine-tuned models to be far from perfect, especially on unseen questions.
We also find that all LLMs we tested, whether fine-tuned or not, are less diverse in their predictions across countries than the actual human survey data.

In summary, we make following \textbf{three main contributions}:
\begin{itemize}[noitemsep]
    \item  We introduce group-level survey response distribution prediction as a simulation task, and share three datasets adapted for training and testing models on this task: two in English and one in Chinese. 
\item We propose a fine-tuning method based on first-token probabilities of multi-choice question answering, and show that this method performs best among the methods tested for our simulation task, which demonstrates the benefits of specialization for simulation. 
\item We contextualize these positive results with evidence of systematic inaccuracies in even the best-performing simulations, thus cautioning against the use of LLMs, specialized or not, for simulating survey response distributions today.%
\footnote{We make all code and used dataset available at \href{https://github.com/yongcaoplus/SimLLMCultureDist}{github.com/yongcaoplus/SimLLMCultureDist}.}
\end{itemize}

\section{Related Work} 

\paragraph{LLM Simulations}
Collecting human response data is one of the most challenging and costly aspects of social science research \cite{argyle2023out,hewitt2024predicting}.
Consequently, much prior work has investigated the extent to which LLMs can accurately simulate human responses in surveys and experimental settings.
Most prominently, \citet{argyle2023out},  \citet{horton2023large} and \citet{aher2023using} all found evidence of LLMs providing reasonably accurate group-level simulations in behavioral science and economics experiments as well as for US political surveys.
Some follow-up work has highlighted biases and conceptual challenges in such simulations \citep{bisbee2023synthetic,bail2024can,park2024diminished,kozlowski2024simulating}.
Other work has explored prompting strategies and frameworks for improving simulation accuracy \citep{kwok2024evaluating,manning2024automated,sun2024random}.
Relatedly, several works have used survey questionnaires, including the WVS used in this paper, with the goal of \textit{evaluating} values and opinions reflected in LLMs rather than simulating human survey responses \citep[][inter alia]{benkler2023assessing,arora-etal-2023-probing,cao-etal-2023-assessing,alkhamissi-etal-2024-investigating,zhao2024worldvaluesbench,wright2024revealingfinegrainedvaluesopinions}.
In contrast to these efforts, our focus is on \textit{specializing} LLMs to simulate group-level survey response distribution, which could aid in survey data collection.
While we do measure the performance of LLMs out of the box (\textit{zero-shot}), our main goal is to investigate the extent to which we can \textit{improve} LLM performance on simulating group-level survey response distributions through fine-tuning, and explore their potential as specialized tools for social science research.


\paragraph{Distribution Simulation as Calibration.}
\textit{Calibration} is aligning classifier predictive probabilities with the classification uncertainty. While most work focuses on majority class accuracy \cite{li-etal-2024-multiple,he2024investigating}, this is problematic when human label variation is substantial \cite{baan-etal-2022-stop,baan-etal-2024-interpreting}, and \textit{human calibration} should consider the full human judgment distribution.
Our task can therefore be viewed as \textit{human calibration} for multiple-choice surveys, as opposed to many previous studies, which focused on accuracy measured against the majority answer \cite{arora-etal-2023-probing, cao-etal-2023-assessing, alkhamissi-etal-2024-investigating}.


\begin{table}[t]
\centering
\resizebox{0.45\textwidth}{!}{%
\small
\begin{tabular}{@{}p{0.11\textwidth} p{0.3\textwidth}@{}}
\toprule
\textbf{Instruction} & How would someone from Andorra answer the following question: \\
\midrule
\textbf{Input} & How interested would you say you are in politics? Here are the options: \\ 
\midrule
\textbf{Options} & \text{(A) Very interested} \\
                 & \text{(B) Somewhat interested} \\
                 & \text{(C) Not very interested} \\
                 & \text{(D) Not at all interested} \\
\midrule
\textbf{Format} & If had to select one of the options, my answer would be ([A/B/...]) \\
\midrule
\textbf{Options} & \text{(A) 15.16\%} \quad \text{(B) 29.02\%}  \\ 
\textbf{Distribution}& \text{(C) 28.31\%} \quad  \text{(D) 27.51\%} \\           \bottomrule
\end{tabular}}
\caption{\textbf{Example entry} from our formatted WVS dataset for the country of Andorra.}
\label{tb:questionnaire_example}
\end{table}

\section{Cultural Survey Simulation Dataset}\label{sec:dataset}

\subsection{Data Source}

We use the 2017-2022 wave of the World Values Survey (WVS)\footnote{\url{https://www.worldvaluessurvey.org/WVSDocumentationWV7.jsp}} to construct our main simulation dataset.
WVS was conducted across 66 countries with over 80,000 respondents. This extensive survey captures societal attitudes on various cultural dimensions, including \textit{family, regional values, education, moral principles, corruption, accountability, etc}.
Our analysis includes all countries with more than 1,000 respondents to ensure robust cross-cultural representation, resulting in a set of 65 countries (Northern Ireland did not qualify).

For our analysis, we use the original questions and answers, excluding validity-check options such as ``not applicable'' and ``refuse to answer'', given their infrequent occurrence in human-collected responses. We conduct experiments using the English and Chinese versions of the datasets obtained from the official source, enabling the analysis of cross-linguistic differences in this task\footnote{We use GLM-4 to translate the missing questions in the Chinese questionnaire.}.

\subsection{Prompt Settings}   

We preserve the original questions and response options, adhering to the GlobalOpinionQA template for consistency \cite{durmus2023globalopinionqa}. As shown in Table~\ref{tb:questionnaire_example}, the model input consists of fields for \textit{instruction}, \textit{input}, \textit{options}, and \textit{format}, while the target for alignment is the \textit{distribution} of the options.  Note that the \textit{format} field is used to restrict the vocabulary of the first token to valid options.

\subsection{Dataset Split} 
We use the first 259 questions of the WVS to construct our dataset, excluding demographics and notes for interviewers. We divide them into three parts based on topics: $Q_1$ (questions 1-163), $Q_2$ (questions 164-198), and $Q_3$ (questions 199-259). Additionally, we divide the countries into three groups to ensure they are challenging to generalize between: $C_1$ (all countries that are not included in the following two sets), $C_2$ (the 8 surveyed countries that are in Africa\footnote{For subset $C_2$, we select countries from one random continent (i.e. Africa). The selected African countries $C_2$ are Egypt, Ethiopia, Kenya, Libya, Morocco, Nigeria, Tunisia, and Zimbabwe. }), and $C_3$ (medium-GDP countries sampled from each continent\footnote{Our selected medium-GDP countries $C_3$ are Malaysia, Thailand, Czechia, Greece, Nigeria, Morocco, Peru, Colombia, Mexico, Puerto Rico, and New Zealand.}).

We split training, validation, and test sets for the aforementioned questions and country subsets. The split and statistical information of the dataset are presented in Table~\ref{tb:dataset_split} and Table~\ref{tb:dataset_statistics} respectively. The test set comprises five subsets, designed to evaluate the performance of models in answering unseen value questions, unseen regional countries, and representative medium-GDP countries.

\subsection{Unseen Survey Dataset}\label{sec:unseen_data} To evaluate generalization to a completely unseen survey, we use an additional subset of GlobalOpinionQA \cite{durmus2023globalopinionqa}, the Pew Global Attitudes Survey (Pew), which maintains a similar format to the WVS but includes different cultural questions. 
We compile two sets of countries for this test set: $C_1^\prime$ and $C_3$. For $C_1^\prime$, we sample ten countries from $C_1$ to maintain geographical and GDP-level diversity and then use the GlobalOpinionQA data specifically for these countries for evaluation (see Appendix~\ref{ax:pew_data_distribution}). For $C_3$, we include the same medium-GDP countries as in $C_3$.

\begin{table}[t]
\small
    \begin{tabular}{p{1.2cm}p{4.8cm}p{0.4cm}}
        \toprule
        \textbf{Countries} & \textbf{Description} & \textbf{N}\\
        \midrule
        \bm{$C_1$} & All WVS countries not in $C_2$ or $C_3$ & 46 \\
        $C_2$ & African countries & 8 \\
        $C_3$ & Medium-GDP countries & 11 \\
        \bottomrule
    \end{tabular}

    \vspace{0.3cm}

    \begin{tabular}{p{1.2cm}p{4.8cm}p{0.4cm}}
        \toprule
        \textbf{Questions} & \textbf{Description} & \textbf{N}\\
        \midrule
        \bm{$Q_1$} & All WVS questions not in $Q_2$ or $Q_3$ & 150 \\
        $Q_2$ & Q's about religious and ethical values & 35 \\
        $Q_3$ & Q's about political interest and culture & 59 \\
        \bottomrule
    \end{tabular}

\caption{\label{tb:dataset_split} 
\textbf{Country and question splits} that we use in our experiments with WVS data. Splits seen during training are highlighted in \textbf{bold}. For additional descriptive statistics on the dataset, see Appendix~\ref{ax:wvs_data_distibution}.
}
\end{table}

\begin{table}[t]
\centering
\resizebox{\columnwidth}{!}{
    \begin{tabular}{l|c|c|c|cc|cc}
        \toprule
        \textbf{Split} & \textbf{Train} & \textbf{Valid} & \multicolumn{5}{c}{\textbf{Test}} \\  \midrule \midrule
        \textbf{Countries} & \bm{$C_1$} & \bm{$C_1$} & \bm{$C_1$} & $C_2$ &  $C_2$ & $C_3$ & $C_3$ \\ 
        \textbf{Questions} &  \bm{$Q_1$} & $Q_2$ & $Q_3$ & \bm{$Q_1$} &  $Q_3$ & \bm{$Q_1$} & $Q_3$ \\  \midrule
        \textbf{Entries} & 6,841 & 1,586 & 2,719 & 1,179 & 471  & 1,644 & 660 \\ 
        \bottomrule
    \end{tabular}}

\caption{
    \label{tb:dataset_statistics} \textbf{WVS dataset statistics} across the country ($C$) and question ($Q$) splits we use in our experiments.
    Splits seen during training are highlighted in \textbf{bold}. Number of entries is not necessarily $C$ $\times$ $Q$ because some entries are missing from survey results.
    }
\end{table}



\section{Methodology}
To address the challenges of simulating culturally diverse survey responses, we introduce a framework for first-token alignment fine-tuning for distribution prediction, designed to improve generalization across populations and survey questions.

\subsection{Probability Distribution Simulation}

Unlike most existing studies that directly prompt LLMs with multiple-choice questions to assess their cultural knowledge or behavior \cite{arora-etal-2023-probing, cao-etal-2023-assessing, alkhamissi-etal-2024-investigating}, we propose a novel task that focuses on simulating the distribution of response options for given questions rather than predicting single answers.

Specifically, let ${Q}$ denote a multiple-choice question and ${O}=\{o_1, o_2, ..., o_n\}$ be the corresponding set of response options, where $n$ is the total number of options, which can vary between questions. The objective is to simulate the option distribution $P(O|{Q})$ to match human response distribution from a particular group (e.g., country).
Consequently, models are evaluated by comparing the alignment of observed vs.\ predicted distributions rather than focusing on majority response options.

\subsection{First-Token Probability Alignment}
Using the dataset introduced in \S\ref{sec:dataset}, we present first-token alignment fine-tuning, to align model outputs with the observed response distributions of specific population groups (e.g., countries).

The processed question $Q$ is used as input into LLMs. The model outputs logits $\{ z_1, z_2, \ldots, z_n \}$ for the first token of each question's corresponding options $O$. The probability distribution for the first token is obtained by applying the softmax function to normalize the indexed option logits:
\[ P_{\text{LLM}}(o_i|{Q}) = \frac{e^{z_i}}{\sum_{j=1}^{n} e^{z_j}} \]

For the training optimization objective, we employ Kullback-Leibler Divergence loss ($\text{Loss}_{\text{KL}}$) to align the LLM's first-token probability distribution with the human response distribution:
\[ \text{Loss}_{\text{KL}} = \sum_{i=1}^{n} P_{\text{human}}(o_i|{Q}) \log \left( \frac{P_{\text{human}}(o_i|{Q})}{P_{\text{LLM}}(o_i|{Q})} \right) \]
where \( P_{\text{human}}(o_i|{Q})\) is the probability of option \( o_i \) based on human survey data, and \( P_{\text{LLM}}(o_i|{Q})\) is the probability output by the LLM.

To improve the efficiency of the fine-tuning process, we implement Low-Rank Adaptation \cite[LoRA;][]{hu2022lora}, a parameter-efficient method specifically designed for optimizing LLMs.



\section{Experimental Setup}

\subsection{Models}
\label{exp:experiment_setup}
We fine-tune seven models across three model families using our processed dataset: Vicuna1.5 \cite{vicuna2023} in its 7B and 13B parameter versions, Llama3 \cite{llama3modelcard} in its 8B Base and Instruct versions, and Deepseek-Distilled-Qwen \cite{guo2025deepseek} in 7B, 14B, and 32B.
Vicuna1.5 is a version of Llama2 \cite{touvron2023llama} fine-tuned on user conversations with ChatGPT, whereas Llama3 is a stronger model. As Vicuna1.5 is less powerful than recent LLMs, it is chosen to evaluate the effect of fine-tuning as an equalizer despite zero-shot performance differences. DeepSeek is a state-of-the-art model series known for its strong performance across diverse benchmarks. We use the DeepSeek-Distilled-Qwen models, which are distilled from the DeepSeek-R1, the current state-of-the-art open weights model.
For further details on all models and our inference setup, see Appendix~\ref{app:hyperparams}.

\begin{table*}[!]
\centering
\resizebox{0.89\textwidth}{!}{
\begin{tabular}{l|l|ccccc|c|ccccc|c}
\toprule
\multirow{2}{*}{Model} & \multirow{2}{*}{Methods} &  \multicolumn{6}{c|}{($1-$JSD) $\uparrow$} & \multicolumn{6}{c}{EMD $\downarrow$} \\  \cmidrule{3-14}
                            & & $C_1$-$Q_3$ & $C_2$-$Q_1$ & $C_2$-$Q_3$ & $C_3$-$Q_1$ & $C_3$-$Q_3$ & \textit{Avg.} & $C_1$-$Q_3$ & $C_2$-$Q_1$ & $C_2$-$Q_3$ & $C_3$-$Q_1$ & $C_3$-$Q_3$ & \textit{Avg.}  \\ \midrule \midrule
\multirow{4}{*}{\textit{Vicuna1.5-7B}} & ZS [ctrl] & 0.732 & 0.754 & 0.748 & 0.761 & 0.747 & 0.748 & 0.095 & 0.108 & 0.095 & 0.106 & 0.086 & 0.098 \\
& ZS & 0.732 & 0.754 & 0.749 & 0.761 & 0.749 & 0.749 & 0.095 & 0.107 & 0.096 & 0.106 & 0.085  &  0.098 \\
& FT [ctrl] & 0.754 & 0.829 & 0.754 & 0.842 & 0.765 & 0.789 & 0.089 & 0.078 & 0.092 & 0.072 & 0.080  & 0.082 \\ 
& \cellcolor{customgray}\textbf{FT} & 
\cellcolor{customgray}\textbf{0.766} & 
\cellcolor{customgray}\textbf{0.859} & 
\cellcolor{customgray}\textbf{0.767} & 
\cellcolor{customgray}\textbf{0.875} & 
\cellcolor{customgray}\textbf{0.775} & 
\cellcolor{customgray}\textbf{0.808} & 
\cellcolor{customgray}\textbf{0.087} & 
\cellcolor{customgray}\textbf{0.071} & 
\cellcolor{customgray}\textbf{0.091} & 
\cellcolor{customgray}\textbf{0.062} & 
\cellcolor{customgray}\textbf{0.078} & 
\cellcolor{customgray}\textbf{0.078}
 \\ \midrule
\multirow{4}{*}{\textit{Vicuna1.5-13B}} & ZS [ctrl]  & 0.735 & 0.755 & 0.743 & 0.775 & 0.747 & 0.751 & 0.095 & 0.102 & 0.096 & 0.096 & 0.086 & 0.095 \\
& ZS & 0.738 & 0.756 & 0.744 & 0.779 & 0.752 & 0.754 & 0.094 & 0.103 & 0.097 & 0.097 & 0.085 & 0.095 \\
& FT [ctrl]  & 0.760 & 0.820 & 0.761 & 0.830 & 0.762 & 0.787 & 0.083 & 0.080 & 0.085 & 0.074 & 0.077 & 0.080 \\ 
& \cellcolor{customgray}\textbf{FT} & 
\cellcolor{customgray}\textbf{0.781} & 
\cellcolor{customgray}\textbf{0.869} & 
\cellcolor{customgray}\textbf{0.781} & 
\cellcolor{customgray}\textbf{0.882} & 
\cellcolor{customgray}\textbf{0.781} & 
\cellcolor{customgray}\textbf{0.819} & 
\cellcolor{customgray}\textbf{0.079} & 
\cellcolor{customgray}\textbf{0.063} & 
\cellcolor{customgray}\textbf{0.084} & 
\cellcolor{customgray}\textbf{0.057} & 
\cellcolor{customgray}\textbf{0.073} & 
\cellcolor{customgray}\textbf{0.071} \\ \midrule
\multirow{4}{*}{\textit{Llama3-8B-Base}} & ZS [ctrl] & 0.748 & 0.766 & 0.757 & 0.779 & 0.768 & 0.764 & 0.097 & 0.114 & 0.093 & 0.109 & 0.088  & 0.100 \\
& ZS & 0.749 & 0.768 & 0.759 & 0.781 & 0.770 & 0.765 & 0.097 & 0.116 & 0.097 & 0.109 & 0.087  & 0.101 \\
& FT [ctrl] & 0.756 & 0.823 & 0.751 & 0.837 & 0.770 & 0.787 & 0.084 & 0.082 & 0.087 & 0.073 & 0.077 & 0.081 \\
& \cellcolor{customgray}\textbf{FT} & 
\cellcolor{customgray}\textbf{0.770} & 
\cellcolor{customgray}\textbf{0.858} & 
\cellcolor{customgray}\textbf{0.773} & 
\cellcolor{customgray}\textbf{0.877} & 
\cellcolor{customgray}\textbf{0.781} & 
\cellcolor{customgray}\textbf{0.812} & 
\cellcolor{customgray}\textbf{0.081} & 
\cellcolor{customgray}\textbf{0.073} & 
\cellcolor{customgray}\textbf{0.087} & 
\cellcolor{customgray}\textbf{0.061} & 
\cellcolor{customgray}\textbf{0.074} & 
\cellcolor{customgray}\textbf{0.075} \\
\midrule
\multirow{4}{*}{\textit{Llama3-8B-Instruct}} & ZS [ctrl] & 0.574 & 0.626 & 0.563 & 0.644 & 0.587 & 0.599 & 0.130 & 0.147 & 0.135 & 0.145 & 0.136  & 0.139 \\
& ZS & 0.585 & 0.650 & 0.589 & 0.657 & 0.584 & 0.613 & 0.130 & 0.139 & 0.126 & 0.145 & 0.141  & 0.136 \\
& FT [ctrl] & 0.756 & 0.826 & 0.751 & 0.833 & 0.764 & 0.786 & 0.077 & 0.077 & 0.082 & 0.073 & 0.071 & 0.076 \\
& \cellcolor{customgray}\textbf{FT} & \cellcolor{customgray}\textbf{0.777} & \cellcolor{customgray}\textbf{0.881} & \cellcolor{customgray}\textbf{0.783} & \cellcolor{customgray}\textbf{0.890} & \cellcolor{customgray}\textbf{0.784} & \cellcolor{customgray}\textbf{0.823} & \cellcolor{customgray}\textbf{0.073}  & \cellcolor{customgray}\textbf{0.060} & \cellcolor{customgray}\textbf{0.080} & \cellcolor{customgray}\textbf{0.053} & \cellcolor{customgray}\textbf{0.067}  & \cellcolor{customgray}\textbf{0.067} \\   
\midrule
\multirow{4}{*}{\textit{Distil-Qwen-7B}} & ZS[ctrl] & 0.586 & 0.641 & 0.642 & 0.698 & 0.682 & 0.650 & 0.084 & 0.096 & 0.138 & 0.100 & 0.123 & 0.108 \\
& ZS & 0.583 & 0.645 & 0.639 & 0.701 & 0.671 & 0.648 & 0.084 & 0.095 & 0.138 & 0.100 & 0.126 & 0.109 \\
& FT[ctrl] & 0.747 & 0.764 & 0.791 & 0.811 & 0.817 & 0.786 & 0.075 & 0.079 & 0.080 & 0.078 & 0.110 & 0.085 \\
& \cellcolor{customgray}\textbf{FT} & 
\cellcolor{customgray}\textbf{0.756} & 
\cellcolor{customgray}\textbf{0.781} & 
\cellcolor{customgray}\textbf{0.803} & 
\cellcolor{customgray}\textbf{0.833} & 
\cellcolor{customgray}\textbf{0.834} & 
\cellcolor{customgray}\textbf{0.801} & 
\cellcolor{customgray}\textbf{0.073} & 
\cellcolor{customgray}\textbf{0.076} & 
\cellcolor{customgray}\textbf{0.077} & 
\cellcolor{customgray}\textbf{0.072} & 
\cellcolor{customgray}\textbf{0.107} & 
\cellcolor{customgray}\textbf{0.081} \\  \midrule
\multirow{4}{*}{\textit{Distil-Qwen-14B}} & 
ZS[ctrl] & 0.650 & 0.704 & 0.749 & 0.711 & 0.743 & 0.712 & 0.088 & 0.082 & 0.089 & 0.115 & 0.141 & 0.103 \\
& ZS  & 0.654 & 0.716 & 0.756 & 0.731 & 0.746 & 0.721 & 0.088 & 0.081 & 0.086 & 0.110 & 0.136 & 0.100 \\
& FT[ctrl] & 0.751 & 0.784 & 0.799 & 0.807 & 0.833 & 0.795 & 0.066 & 0.074 & 0.084 & 0.083 & 0.103 & 0.082 \\
& \cellcolor{customgray}\textbf{FT} & 
\cellcolor{customgray}\textbf{0.777} & 
\cellcolor{customgray}\textbf{0.816} & 
\cellcolor{customgray}\textbf{0.827} & 
\cellcolor{customgray}\textbf{0.851} & 
\cellcolor{customgray}\textbf{0.861} & 
\cellcolor{customgray}\textbf{0.826} & 
\cellcolor{customgray}\textbf{0.061} & 
\cellcolor{customgray}\textbf{0.068} & 
\cellcolor{customgray}\textbf{0.077} & 
\cellcolor{customgray}\textbf{0.070} & 
\cellcolor{customgray}\textbf{0.099} & 
\cellcolor{customgray}\textbf{0.075} \\ \midrule
\multirow{4}{*}{\textit{Distil-Qwen-32B}} & ZS[ctrl]  & 0.649 & 0.671 & 0.721 & 0.674 & 0.743 & 0.691 & 0.089 & 0.105 & 0.099 & 0.131 & 0.126 & 0.110 \\
& ZS   & 0.658 & 0.683 & 0.719 & 0.695 & 0.749 & 0.701 & 0.088 & 0.105 & 0.101 & 0.129 & 0.125 & 0.110 \\
& FT[ctrl]  & 0.776 & 0.789 & 0.799 & 0.794 & 0.821 & 0.796 & 0.066 & 0.071 & 0.074 & 0.087 & 0.111 & 0.082 \\
& \cellcolor{customgray}\textbf{FT} & 
\cellcolor{customgray}\textbf{0.800} & 
\cellcolor{customgray}\textbf{0.821} & 
\cellcolor{customgray}\textbf{0.815} & 
\cellcolor{customgray}\textbf{0.830} & 
\cellcolor{customgray}\textbf{0.846} & 
\cellcolor{customgray}\textbf{0.822} & 
\cellcolor{customgray}\textbf{0.062} & 
\cellcolor{customgray}\textbf{0.067} & 
\cellcolor{customgray}\textbf{0.073} & 
\cellcolor{customgray}\textbf{0.078} & 
\cellcolor{customgray}\textbf{0.110} & 
\cellcolor{customgray}\textbf{0.078}  \\
\bottomrule
\end{tabular}}
\caption{\label{tb:main_results}\textbf{Main results for predicting country-level survey response distributions on the WVS data}.
We test all models with zero-shot prompting (ZS) and our proposed fine-tuning approach (FT). [ctrl] indicates a control setup, where we randomly replace countries in test prompts with other countries, to evaluate country context sensitivity.
We report Jensen-Shannon Divergence ($1-$JSD , $\uparrow$) and Earth Mover Distance (EMD, $\downarrow$).}
\end{table*}

\subsection{Baselines}
In the main body of this paper, we compare our fine-tuning method (FT) to a zero-shot prompting (ZS) baseline, which is the default method explored in prior work.
ZS involves directly querying the models with the country context and questions.
As an additional control setting, for both ZS and FT, we replace countries in the queries with other countries randomly selected from among the full set of countries in the WVS. This approach is designed to investigate the sensitivity of the LLMs to the specific country given in the context vs.\ the prior distributions of response options.
We denote this control setting as ``[ctrl]''.
In Appendix \ref{ax:more_baseline}, we show additional baselines such as K-Nearest Neighbors, which generally perform worse than FT.

\subsection{Metrics}
To measure the alignment of predicted response distributions with country-level reference distributions, we adopt two metrics for evaluation: i) \textbf{1-JSD}, where JSD is Jensen-Shannon divergence, also employed by \citet{durmus2023globalopinionqa}, is a symmetric measure of the similarity between two probability distributions, with higher values indicating greater similarity; ii) \textbf{Earth Mover Distance} \cite[EMD;][]{rubner1998metric}, also known as the Wasserstein distance, quantifies the minimum amount of work required to transform one distribution into another, with lower values indicating greater similarity in distribution. Both the 1-JSD and EMD metrics range from 0 to 1.

\section{Results}
We investigate two primary research questions: 
\begin{description} 
\item[RQ1] How effectively does the proposed alignment method improve the distribution simulation of the model on unseen countries and questions?
\item[RQ2] What is necessary to perform well on the task---how much is dependent on modeling the prior distribution, and how much on context sensitivity?
\end{description}
We present comprehensive experimental results to address the two research questions.

\subsection{RQ1: Generalization Performance}
To address RQ1 (how effectively FT improves distribution simulation on unseen countries and questions), we train the selected models using our proposed simulation methods and evaluate their performance on unseen countries and questions. Table~\ref{tb:main_results} presents the evaluation scores across models of varying sizes and types.

\paragraph{Zero-shot [ZS] vs.\ Fine-tuning [FT].} Across all model sizes and types, we observe that zero-shot prompting (ZS) consistently yields worse scores compared to fine-tuned models (FT), indicating that while ZS is capable of addressing unseen countries and questions, it lacks the adaptability needed for effective distribution simulation. In contrast, fine-tuned models show improved performance, with higher $1-$JSD and lower EMD scores (e.g., a \textit{34.3\% $1-$JSD} increase and \textit{0.069 EMD} decrease for Llama3-8B-Instruct \textit{Avg.}), demonstrating their enhanced ability to align with real-world distributions. This suggests that fine-tuning enables models to internalize more detailed patterns and relationships, making them more effective for simulation.

\paragraph{Unseen Countries vs.\ Unseen Questions.} The generalization capabilities are revealed by evaluation on unseen attributes. Across all models and settings, unseen questions ($Q_3$) tend to present a greater challenge than unseen countries ($C_2$ or $C_3$), as indicated by slightly worse scores for unseen questions (e.g., \textit{0.781} vs.\ \textit{0.886 $1-$JSD} for Llama3-Instruct). This suggests that while the models are reasonably robust in handling new country distributions, they struggle more with generating accurate distributions for questions that were not encountered during training. 

\paragraph{Comparison Across Models.} 
Our method demonstrates consistent performance improvements across three representative model families and various model sizes, highlighting the effectiveness of our proposed first-token alignment approach. Specifically, while there are notable performance differences in the zero-shot setting—for example, Llama3-Base achieves a higher $1-$JSD value (0.765) compared to Llama3-Instruct (0.613), suggesting that the base model better predicts the option token distribution—our fine-tuning procedure significantly bridges this gap. After fine-tuning, Llama3-Instruct not only closes the initial performance disparity but even surpasses Llama3-Base (\textit{34.3\% $1-$JSD increase} vs. \textit{6.1\%} increase). Similarly, experiments with Distilled Qwen models of different sizes reveal no clear scaling trends. Moreover, although Vicuna1.5 is generally considered weaker compared to Llama and Qwen, it surprisingly delivers similarly competitive results on this task. Overall, analysis across all models further revealing the finding that, \textbf{regardless of the starting model, our fine-tuning approach consistently produces models with similar strong performance on our task}.

\begin{figure}[t]
    \centering
    \includegraphics[width=0.48\textwidth]{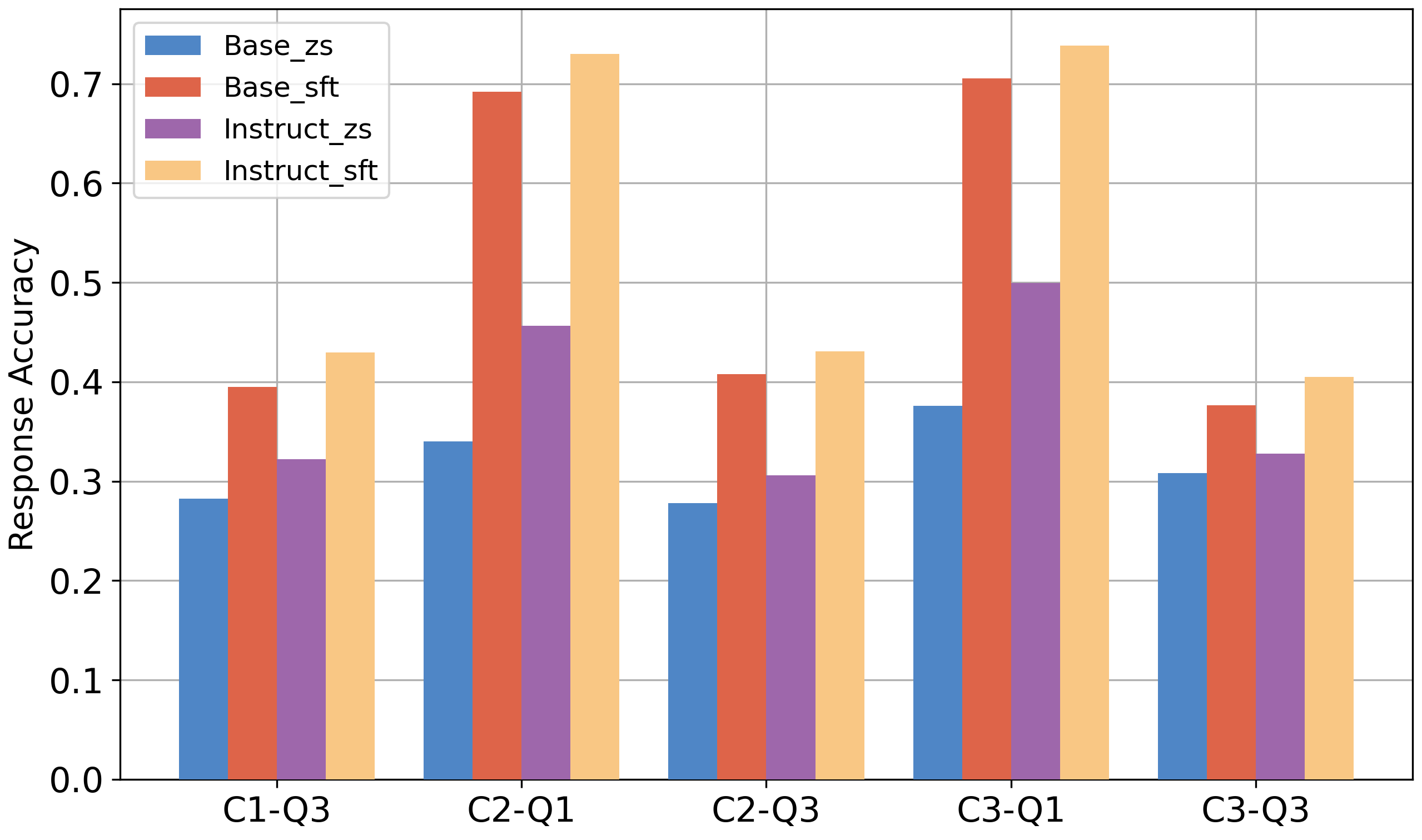}
    \caption{Option prediction accuracy for cultural questions using the Llama3-8B-Instruct. The final option is simulated by selecting the option with the highest probabilities,  compared against human majority choice.}
    \label{fig:prediction_acc}
\end{figure}

\begin{figure*}[t]
    \centering
    \includegraphics[width=0.99\textwidth]{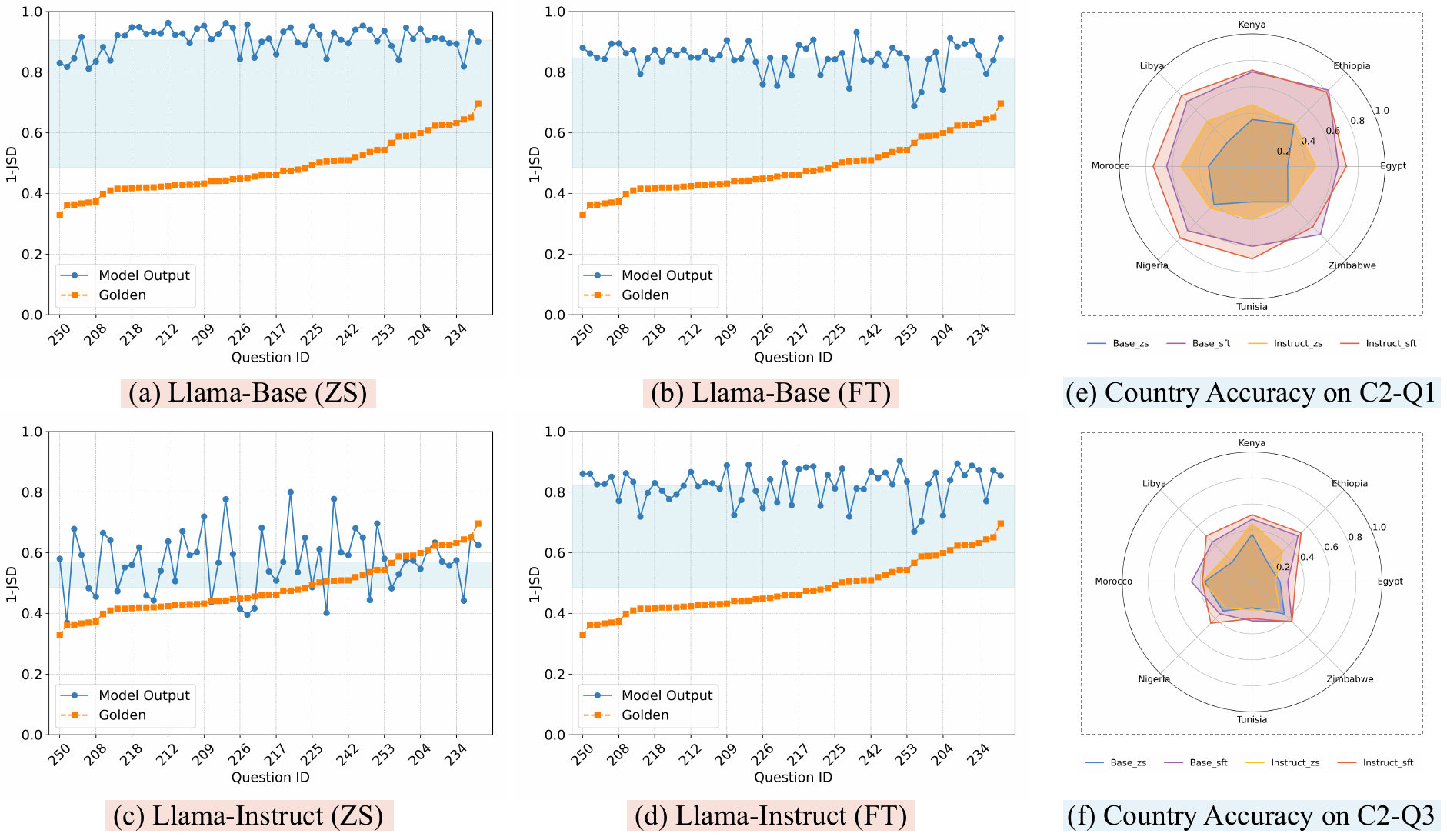} 
    \caption{Model Diversity and Country Accuracy Analysis. (a)-(d) denotes the comparison of $1-$JSD scores across countries for specific questions (\textit{$C_2$-$Q_3$} and \textit{$C_3$-$Q_3$}). The blue-shaded area represents diversity changes, with the lower boundary indicating the mean of survey response scores and the upper boundary representing the mean of model outputs. (e)-(f) denotes the accuracy of options on African countries in both \textit{$C_2$-$Q_1$} and \textit{$C_2$-$Q_3$}.}
    \label{fig:diversity_distribution}
\end{figure*}

\paragraph{Correlation Between First-Token Distribution and Response Accuracy.} Beyond the alignment of first-token distributions, we further evaluate of the mode accuracy of models in responding to questionnaire items following fine-tuning. We consider the options with the highest probabilities as (single, argmax) predictions and calculate accuracy against the survey majority choice per country \cite{arora-etal-2023-probing, cao-etal-2023-assessing, alkhamissi-etal-2024-investigating}. As depicted in Figure~\ref{fig:prediction_acc}, our findings reveal a significant increase in accuracy across all models and test subsets, highlighting the effectiveness of the fine-tuning process. Notably, performance improvements are particularly significant in unseen countries, as demonstrated by $C_2$-$Q_1$ and $C_3$-$Q_1$ subsets. These results suggest that our proposed method not only improves the simulation of option distributions but also strengthens the models' alignment with the correct responses, underscoring the interdependence of distribution alignment and answer accuracy.

\begin{table*}[!]
\centering
\resizebox{0.999\textwidth}{!}{
\begin{tabular}{l|c|ccccc|c|ccccc|c}
\toprule
\multirow{2}{*}{Base Model} & \multirow{2}{*}{Methods} &  \multicolumn{6}{c|}{($1-$JSD) $\uparrow$} & \multicolumn{6}{c}{EMD $\downarrow$} \\  \cmidrule{3-14}
                            & & $C_1$-$Q_3$ & $C_2$-$Q_1$ & $C_2$-$Q_3$ & $C_3$-$Q_1$ & $C_3$-$Q_3$ & \textit{Avg.} & $C_1$-$Q_3$ & $C_2$-$Q_1$ & $C_2$-$Q_3$ & $C_3$-$Q_1$ & $C_3$-$Q_3$ & \textit{Avg.}  \\ \midrule \midrule 
\multirow{2}{*}{\textit{Llama3-8B-Instruct}} & ZS & 0.603 & 0.654 & 0.601 & 0.659 & 0.613 & 0.626 & 0.125 & 0.140 & 0.131 & 0.145 & 0.132 & 0.135 \\
& \cellcolor{customgray}\textbf{FT} & 
\cellcolor{customgray}\textbf{0.777} & 
\cellcolor{customgray}\textbf{0.852} & 
\cellcolor{customgray}\textbf{0.789} & 
\cellcolor{customgray}\textbf{0.870} & 
\cellcolor{customgray}\textbf{0.791} & 
\cellcolor{customgray}\textbf{0.816} & 
\cellcolor{customgray}\textbf{0.081} & 
\cellcolor{customgray}\textbf{0.078} & 
\cellcolor{customgray}\textbf{0.087} & 
\cellcolor{customgray}\textbf{0.065} & 
\cellcolor{customgray}\textbf{0.073} & 
\cellcolor{customgray}\textbf{0.077} \\ \midrule
\multirow{2}{*}{\textit{Distil-Qwen-7B}} &
ZS & 0.605 & 0.706 & 0.712 & 0.696 & 0.693 & 0.682 & 0.084 & 0.091 & 0.086 & 0.087 & 0.190 & 0.108 \\
& \cellcolor{customgray}\textbf{FT} & 
\cellcolor{customgray}\textbf{0.764} & 
\cellcolor{customgray}\textbf{0.779} & 
\cellcolor{customgray}\textbf{0.771} & 
\cellcolor{customgray}\textbf{0.791} & 
\cellcolor{customgray}\textbf{0.851} & 
\cellcolor{customgray}\textbf{0.791} & 
\cellcolor{customgray}\textbf{0.067} & 
\cellcolor{customgray}\textbf{0.085} & 
\cellcolor{customgray}\textbf{0.081} & 
\cellcolor{customgray}\textbf{0.082} & 
\cellcolor{customgray}\textbf{0.130} & 
\cellcolor{customgray}\textbf{0.089} \\
\bottomrule
\end{tabular}}
\caption{\label{tb:zh_results}\textbf{Model Performance on Chinese World Values Survey}. 
Performance comparison between zero-shot prompting (ZS) and supervised fine-tuning (SFT) on two model families.}
\end{table*}

\subsection{RQ2: Variation Sensitivity}
To address RQ2 (contribution of modeling the prior distribution vs.\ context sensitivity), we compare the control setting (with countries randomly replaced), analyze diversity changes of models, and explore shifts in response accuracy for unseen countries.

\paragraph{ZS[ctrl] vs. FT[ctrl].} In the control setting, ZS[ctrl] shows a smaller performance drop than ZS, while FT[ctrl] sees a larger decline, with a \textit{16.7\% avg. ($1-$JSD)} drop between FT[ctrl] and FT across seven models, compared to \textit{3.7\%} for ZS[ctrl] and ZS. This indicates that fine-tuned models (FT) are more sensitive to the country context and not just the prior distribution of responses (which FT[ctrl] is trained to simulate), suggesting they have become more specialized in capturing cultural nuances during training. In contrast, the smaller gap between ZS and ZS[ctrl] implies that zero-shot models maintain a more generalized understanding of cultural contexts, making them less affected by random permutations of country data. This difference highlights the fine-tuned models’ improved capability in simulating response distributions for specific countries.


\paragraph{Country Diversity in Model Outputs.}
We define diversity as the divergence across countries of response distributions given the same question. To assess whether FT can enhance model output diversity, we calculate $1-$JSD for Llama3-Base and Llama3-Instruct (ZS and FT) output across countries for each question.  A lower $1-$JSD score suggests greater diversity in responses between countries, whereas a higher score indicates greater similarity in distributions. Results are shown in Figure~\ref{fig:diversity_distribution}, where the scores of survey responses are compared with those of the model predictions.

This visualization provides several interesting insights. Firstly, the outputs of Base models exhibit a high degree of uniformity across countries, indicating a \textbf{limited sensitivity to national variations} when addressing cultural values questions. After fine-tuning, there is a slight reduction in $1-$JSD, suggesting an enhancement in the responsiveness to diverse cultural contexts. Secondly, although the Instruct model, having undergone alignment fine-tuning, initially produces a more varied distribution of answers, \textbf{this diversity diminishes following distribution simulation fine-tuning}. Thirdly,
we observe \textbf{no consistent correlation between the diversity of generated responses and the accuracy of simulated distributions}. Lastly, the post-fine-tuning diversity of responses from both models converges, indicating that our fine-tuning approach \textbf{improves the sensitivity to national differences}.

\begin{table}[t]
\centering
\resizebox{0.98\columnwidth}{!}{
\begin{tabular}{l|cc|cc}
\toprule
\multirow{2}{*}{Methods}  & \multicolumn{2}{c|}{$C_1^\prime$} & \multicolumn{2}{c}{$C_3$}
 \\ \cmidrule{2-5}
& ($1-$JSD) $\uparrow$ & ACC $\uparrow$ & ($1-$JSD) $\uparrow$ & ACC $\uparrow$   \\ \midrule \midrule
\textit{Vicuna1.5} (ZS)         & 0.690 & 0.360  & 0.668 & 0.346 \\
\textit{Vicuna1.5} (FT)        & \cellcolor{customgray}\textbf{0.725} & 
\cellcolor{customgray}\textbf{0.442} & 
\cellcolor{customgray}\textbf{0.709} &  
\cellcolor{customgray}\textbf{0.456} \\ \midrule
\textit{Llama3} (ZS)  & 0.617 & 0.472 &   0.613    &   0.446    \\
\textit{Llama3} (FT) & \cellcolor{customgray}\textbf{0.767} & 
\cellcolor{customgray}\textbf{0.562} &  
\cellcolor{customgray}\textbf{0.755} &  
\cellcolor{customgray}\textbf{0.568} \\ \bottomrule  
\end{tabular}}
\caption{\label{tb:pew_evaluation} Evaluation results on GlobalOpinionQA Pew dataset for \textit{Llama3-Instruct} and \textit{Vicuna1.5-7B}. ($1-$JSD) and option accuracy scores are reported. }
\end{table}

\paragraph{Unseen Country Shifts.} 
We visualize the option accuracy of African countries as observation objects in Figure~\ref{fig:diversity_distribution}e-\ref{fig:diversity_distribution}f. For seen questions, we find that all models show a relatively high performance across most countries, with Ethiopia and Nigeria displaying the highest accuracies close to 80\%. For unseen questions, the performance decreases greatly compared to seen questions, particularly in countries like Egypt and Tunisia. Besides, the Instruct models maintain a higher level of accuracy relative to the Base models in most cases, showing the relative robustness of Instruct models in both seen and unseen scenarios across the African countries, which is consistent with results in Figure~\ref{fig:prediction_acc}.

\subsection{Robustness Analysis}
In this section, we explore the robustness of the models on survey language and unseen survey. 

\paragraph{Impact of Survey Language on Results.}
We fine-tuned the Llama3-8B-Instruct and Distil-Qwen-7B models on the official Chinese translation dataset, and the results are presented in Table \ref{tb:zh_results}. While the performance of both models in Chinese is marginally lower than in English both on (1-JSD) and EMD metrics, the difference is not significant, suggesting that current LLMs exhibit limited sensitivity to language differences in this task. Besides, while the Distilled-Qwen model demonstrates a better performance than Llama3 on most benchmarks, it does not outperform Llama3 in this task. 

\paragraph{Generalization to a New Survey.}
We use Pew introduced in \S\ref{sec:unseen_data} to test the generalization of our fine-tuned models.
Table~\ref{tb:pew_evaluation} presents the results in both $1-$JSD scores and accuracy. Notably, all metric scores show significant improvements for both models after fine-tuning. Additionally, across both $C_1^\prime$ and $C_3$, Llama3 outperforms Vicuna1.5 in terms of accuracy, particularly in the fine-tuned setting, where Llama3 (FT) achieves \textit{19.1\%} and \textit{27.4\%} improvement for two datasets, respectively. The consistent improvements prove its capability to generalize well to unseen surveys.

\begin{table}[t]
\centering
\resizebox{0.98\columnwidth}{!}{
\begin{tabular}{l|ccccc}
\toprule
Methods         & $C_1$-$Q_3$ & $C_2$-$Q_1$ & $C_2$-$Q_3$ & $C_3$-$Q_1$ & $C_3$-$Q_3$ \\ \midrule \midrule
\cellcolor{customgray}\textbf{KL (Orig)}   & \cellcolor{customgray}\textbf{0.777} & 
\cellcolor{customgray}\textbf{0.881} & 
\cellcolor{customgray}\textbf{0.783} & 
\cellcolor{customgray}\textbf{0.890} & 
\cellcolor{customgray}\textbf{0.784}  \\
\midrule
WA Loss &   0.733 & 0.774 & 0.744 & 0.782 & 0.749 \\
JS Loss   & 0.745 & 0.790 & 0.756 & 0.799 & 0.763   \\
CE Loss & 0.746 & 0.809 & 0.772 & 0.807 & 0.756 \\
 \midrule
Shuffled  &  0.753 & 0.820 & 0.761 & 0.815 & 0.761    \\
\bottomrule
\end{tabular}}
\caption{\label{tb:ablation_study} \textbf{Results of our ablation studies}.
We compare different loss functions (WA, JS and CE) to our KL loss setup.
We also evaluate on a test set with shuffled option orders.
All results are for Llama3-8B-Instruct.}
\end{table}

\subsection{Ablation Studies}
We conduct ablation studies to analyze the impact of the training loss function and option ordering.

\paragraph{Loss Function.} As shown in Table~\ref{tb:ablation_study}, KL Loss is the most effective loss function for our task. However, Wasserstein (WA) Loss Jensen-Shannon (JS), and Cross-Entropy (CE) Loss also improve over zero-shot prompting.

\paragraph{Option Ordering.} \citet{dominguez2023questioning} observed an A-bias effect, where models tend to disproportionately select the answer choice labeled ``A''. To assess this bias, we re-evaluate our fine-tuned model on the same dataset where the answer options are shuffled. As shown in Table~\ref{tb:ablation_study} (``Shuffled''), there is a reduction in performance, but it is small compared to the effect of fine-tuning or model choice. This indicates that the option ordering is not a major concern in our experiments.


\section{Conclusion}
In this paper, we explored the task of specialising LLMs to simulate survey response distributions across diverse countries and questions.
For this task, we devised a fine-tuning method based on first-token probabilities.
Our experiments demonstrate that fine-tuning models substantially improves response simulation prediction compared to zero-shot models, for both seen and unseen countries and questions. Further, fine-tuned models also show improved generalization to an entirely new survey dataset. Despite these improvements, our results also highlight systematic limitations of the models, particularly when simulating responses to unseen questions. We also observed that the models, whether fine-tuned or not, were less diverse in their predictions compared to the human survey response data, raising questions about their utility.

While our results provide clear evidence for the benefits of specializing LLMs for survey simulation tasks, they also underscore the need for caution when using LLMs for this task, as even the best-performing models exhibited systematic inaccuracies, especially in culturally diverse contexts.


\section*{Limitations} 
While we proved the effectiveness of our proposed method, several limitations remain in our work.

\paragraph{Scope.} Our trained models are highly specialized and can only be used for the specific task of predicting the distribution of answers to a given survey question from a specified human population. Future work will investigate whether the fine-tuning approach also results in less biased or more aligned models in general-purpose applications, but this cannot be claimed only based on our study.

\paragraph{Language and Countries Coverage.} Our study only uses English prompts for experiments and uses countries to represent specific cultures, consistent with existing studies \cite{cao-etal-2023-assessing, alkhamissi-etal-2024-investigating}. While this approach offers some valuable insights, it may limit the applicability of our findings to non-English LLMs and diverse fine-grained cultural contexts. We hope that future research could benefit from exploring broader languages and countries to enhance the robustness of the proposed framework.

\paragraph{Model Choice.} Due to computational resource consideration, we did not fine-tune LLMs with more than 32B parameters and instead selected a limited number of models for validation. Despite this limitation, in future work, we aim to cover a range of powerful models of varying sizes, which will allow us to uncover interesting observations regarding their performance. We believe that the insights we draw will still contribute to future research, encouraging further exploration of larger models to better understand their capabilities in simulating cultural diversity.


\section*{Ethics Statement}
This research adheres to strict ethical standards, ensuring that all datasets, large language models, and prompt settings used are sourced from open-access repositories and are properly licensed to their original creators. 

While our proposed framework does not involve any inherently risky operations, we acknowledge that the deployment of LLMs carries inevitable potential ethical implications. Therefore, users interacting with our models are strongly encouraged to consider safety and ethical factors, remaining aware of the potential risks and harms that may arise from misuse or misinterpretation of the generated content. Through this work, we aim to contribute positively to a better understanding of cultural diversities and promote responsible practices in the simulation of cultural contexts.

\section*{Acknowledgments}
We thank anonymous reviewers for their valuable
comments. This research was co-funded by a DFF Sapere
Aude research leader grant under grant agreement
No 0171-00034B, and supported by the Pioneer Centre for AI, DNRF grant number P1.  
Yong Cao was supported by a VolkswagenStiftung Momentum grant.
\bibliography{anthology, custom, pr_refs}

\appendix

\section{Pew Data Distribution}
\label{ax:pew_data_distribution}
The detailed statistics for the C2 and C3 split of the cultural survey simulation and the C3 split of GlobalOpinionQA data mentioned are available in Table~\ref{tb:c2_c3}.  The sampled countries, along with the sampling reason of Pew-$C_1^\prime$, are in Table~\ref{tb:pew-c1}.

\begin{table}[ht]
\small
\begin{tabular}{ll|lll}
\toprule
\multicolumn{2}{c|}{C2} & \multicolumn{3}{c}{C3}  \\ \midrule \midrule
Country       & CSS-N      & Country     & CSS-N & PWE-N \\
Egypt         &  157      & Malaysia    & 150  & 516      \\
Ethiopia      &  185      & Thailand    & 150  & 319      \\
Kenya         &  185      & Czechia     & 150  & 212      \\
Libya         &   185     & Greece      & 150  & 648      \\
Morocco       &   184     & Nigeria     & 210  & 1044      \\
Nigeria       &   185     & Morocco     & 209  & 450      \\
Tunisia       &   184     & Peru        & 146  & 545      \\
Zimbabwe      &   185     & Colombia    & 150  & 369      \\
              &        & Mexico      &  150 & 890      \\
              &        & Puerto Rico &  150 & 221      \\
              &        & New Zealand &  149 & 274      \\ \bottomrule
\end{tabular}
\caption{Number of Entries Used in C2 and C3 Split of Cultural Survey Simulation and C3 Split of GlobalOpinionQA Data.}
\label{tb:c2_c3}
\end{table}

\begin{table}[ht]
\small
\begin{tabular}{llll}
\toprule
\textbf{Country} & \textbf{Continent} & \textbf{GDP-level} & \textbf{N} \\ \midrule \midrule
Germany         & Europe            & High               & 1130             \\
Japan            & Asia              & High               & 891              \\
Brazil          & South America     & Upper-middle       & 922              \\
Australia        & Oceania           & High               & 627              \\
India            & Asia              & Lower-middle       & 932              \\
Nigeria          & Africa            & Lower-middle       & 1044             \\
United States    & North America     & High               & 1104             \\
Vietnam          & Asia              & Lower-middle       & 471              \\
Chile            & South America     & Upper-middle       & 542              \\
Ukraine          & Europe            & Lower-middle       & 661              \\ \bottomrule
\end{tabular}
\caption{Sampled countries for Pew-$C_1^\prime$ regarding geographical and GDP-level diversity. The dataset consists of 8,324 entries in total.\label{tb:pew-c1}}
\end{table}

\section{WVS data Distribution}
\label{ax:wvs_data_distibution}
\begin{figure*}[t]
    \centering
    \includegraphics[width=0.98\textwidth]{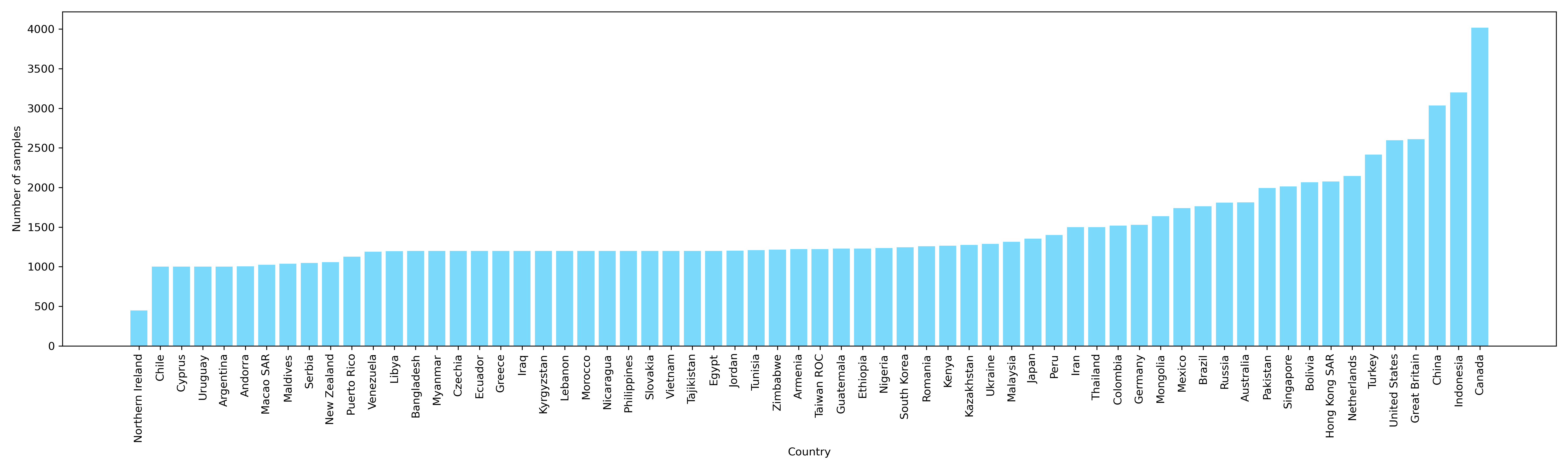} 
    \caption{Sample distribution of the World Values Survey across countries, including only countries with more than 1,000 samples for balanced representation.}
    \label{fig:wvs_sample_distribution}
\end{figure*}

Figure \ref{fig:wvs_sample_distribution} illustrates the sample distribution across various countries in World Values Survey. For the purpose of balanced representation, in our formatted dataset, we only include countries with more than 1,000 samples, ensuring consistency in data distribution across countries and minimizing sample size discrepancies in the analysis.

\begin{figure*}[t]
    \centering
    \begin{subfigure}[b]{0.7\textwidth}
        \centering
        \includegraphics[width=\textwidth]{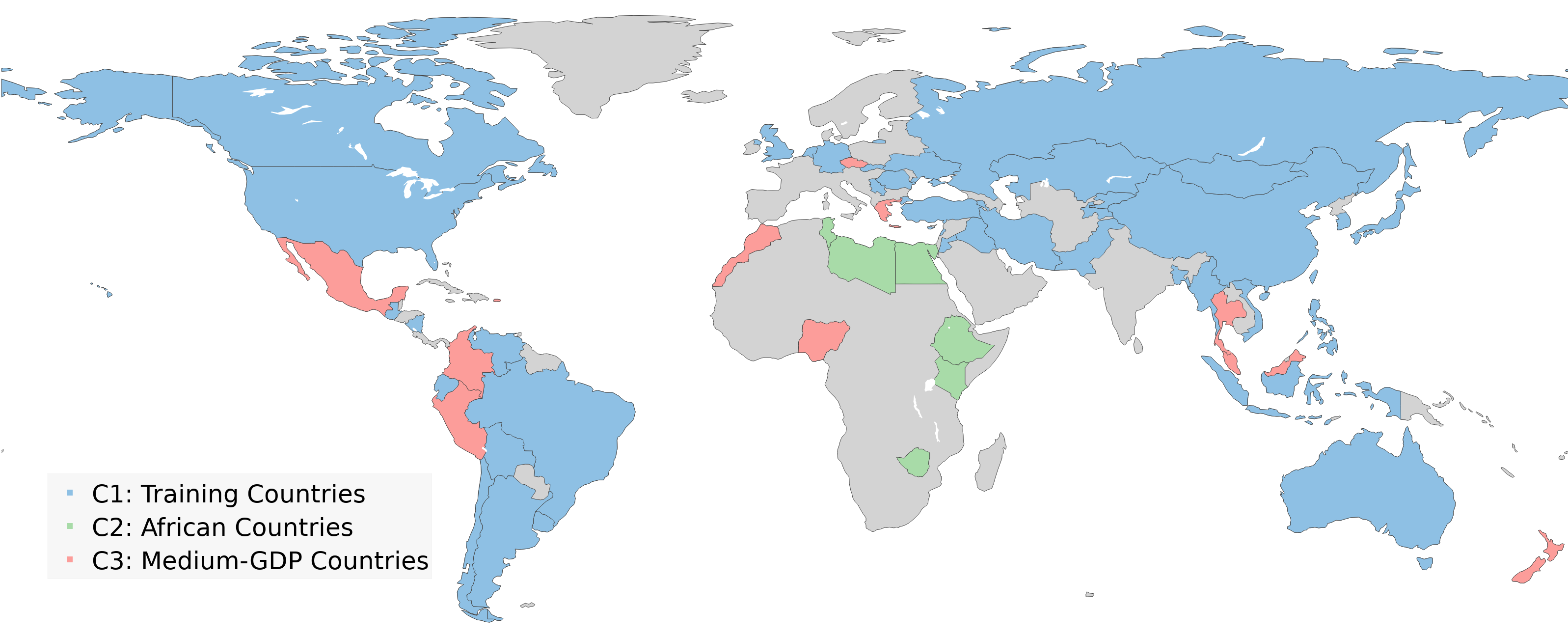}
        \caption{Country distribution across training, validation, and testing sets.}  
        \label{fig:country_split}
    \end{subfigure}
    \hfill
    \begin{subfigure}[b]{0.29\textwidth}
        \centering
        \includegraphics[width=\textwidth]{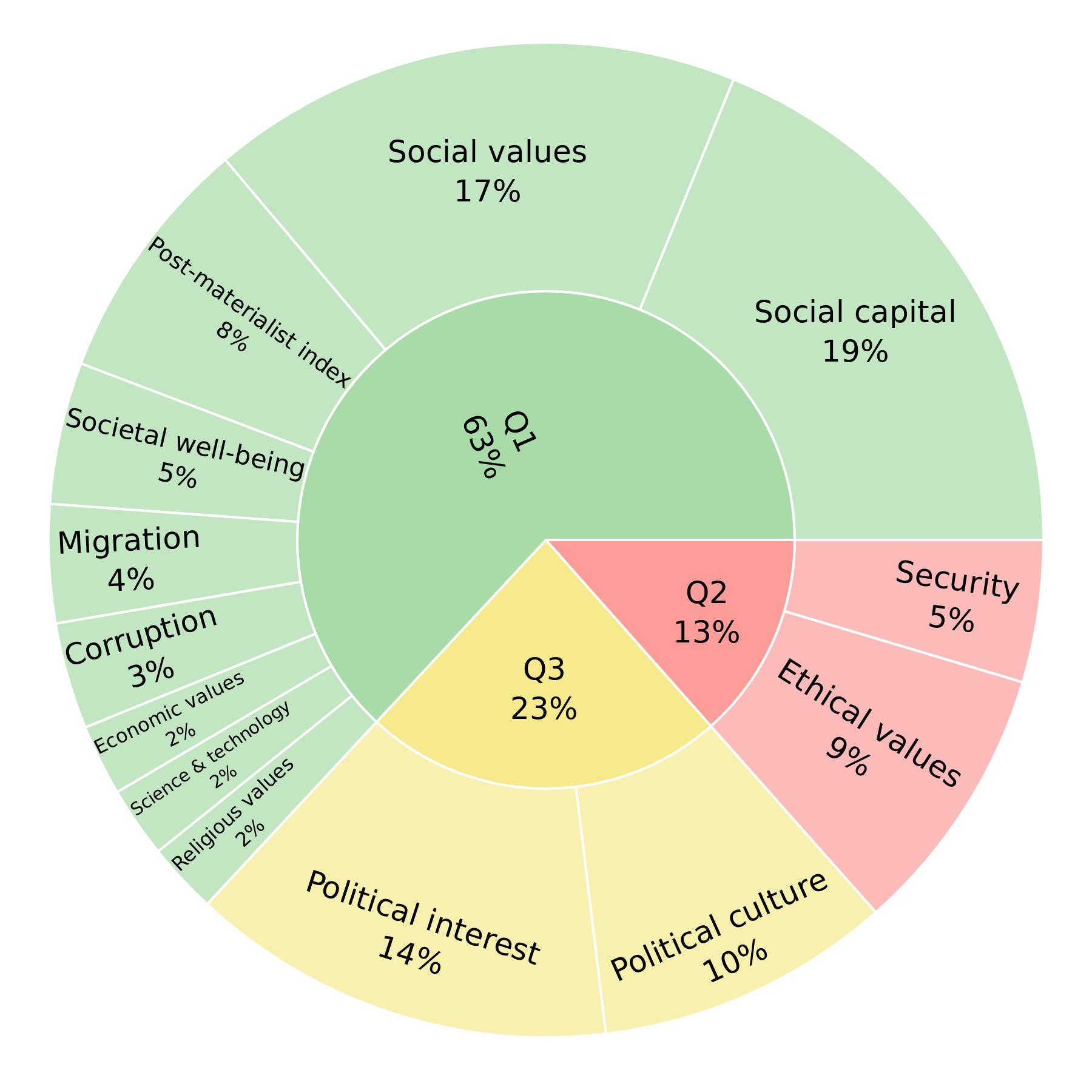}
        \caption{Cultural dimension distribution.}  
        \label{fig:question_split}
    \end{subfigure}
    \caption{Visualization of country and cultural dimension divisions of WVS. Countries are categorized into three groups, and questions are divided based on selected cultural dimensions.}
    \label{fig:wvs_distritbuion}
\end{figure*}

Regarding dataset split, we split WVS by cultural dimensions and countries vis multiple sets, which is visualized in Figure~\ref{fig:wvs_distritbuion} and the corresponding cultural dimensions are detailed in Table~\ref{tb:survey_dimensions}, offering a comprehensive view of the overall distribution in the cultural survey simulation.

\begin{figure}[ht]
    \centering
    \begin{subfigure}[b]{0.46\textwidth}
        \centering
        \includegraphics[width=\textwidth]{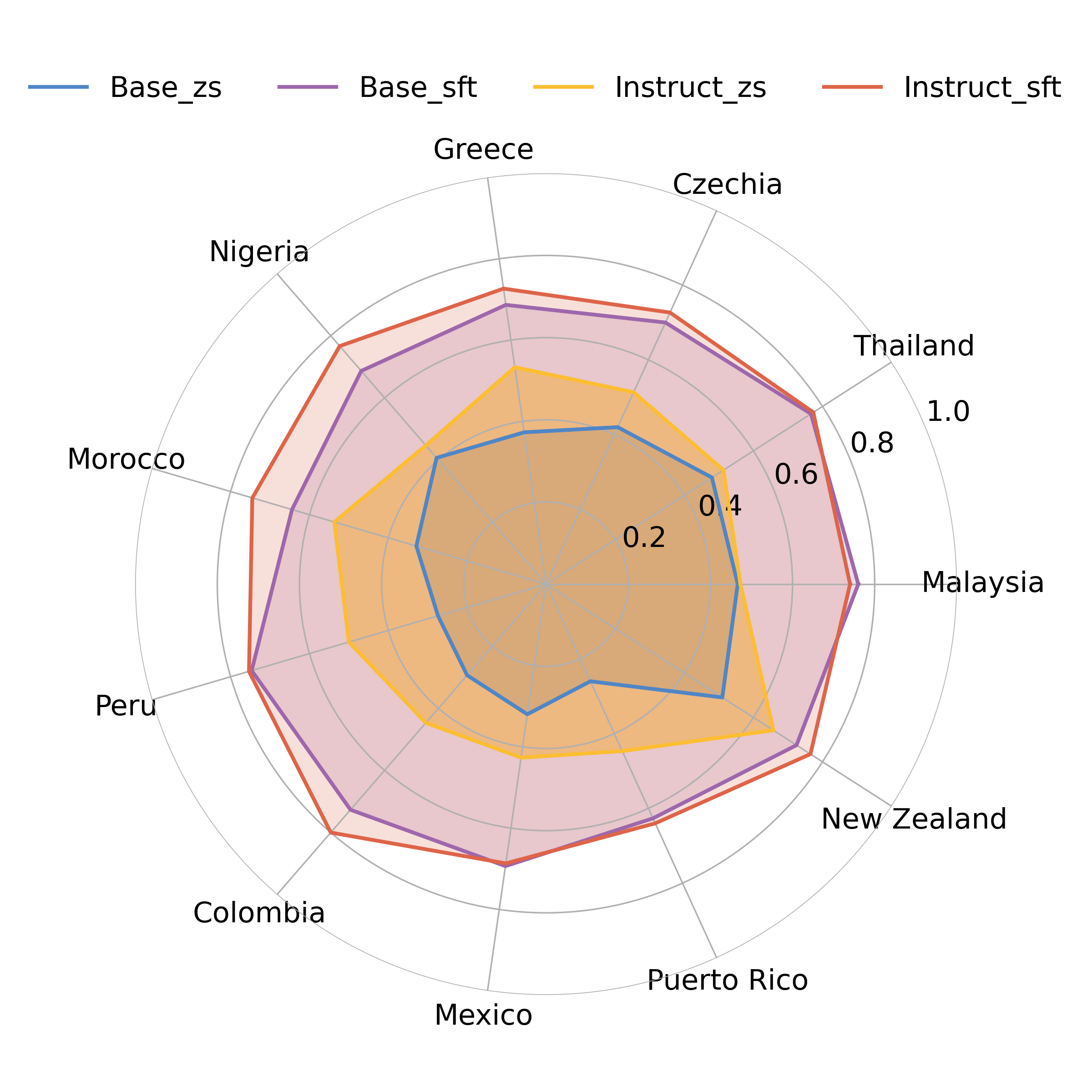}
        \caption{Country Accuracy on C3-Q1.}  
        \label{fig:country_accuracy}
    \end{subfigure}
    \hfill
    \begin{subfigure}[b]{0.46\textwidth}
        \centering
        \includegraphics[width=\textwidth]{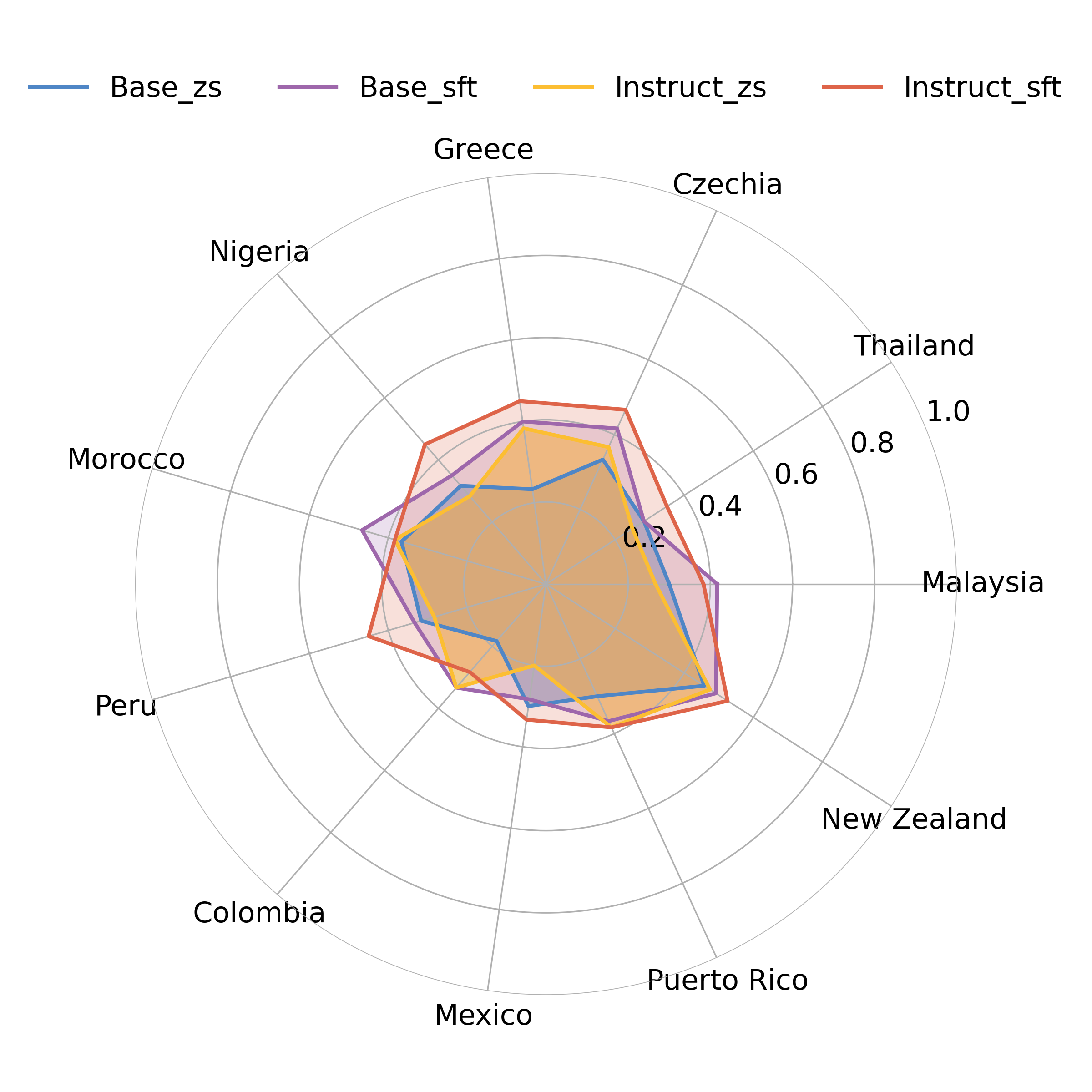}
        \caption{Country Accuracy on C3-Q3.}  
        \label{fig:question_accuracy}
    \end{subfigure}
    \caption{Llama3 Accuracy of options on African countries in both seen C3-Q1 and unseen C3-Q3 questions.}
    \label{fig:acc_wvs_c3}
\end{figure}

\begin{table}[t]
\centering
\resizebox{\columnwidth}{!}{
\begin{tabular}{c|p{8.5cm}c}
\toprule
\textbf{Set} & \textbf{Cultural Dimension} & \textbf{Number} \\ 
\midrule
\multirow{9}{*}{$Q_1$} & Social values, attitudes and stereotypes    & 45  \\
 & Societal well-being       & 12   \\
 & Social capital, trust and organizational membership &  49  \\
 & Economic values                                    &   6   \\
 & Corruption                                          &   9   \\
 & Migration                                           &  10    \\
 & Post-materialist index                              &  21   \\
 & Science  and technology                                & 6   \\
 & Religious values                                    &    6   \\ \midrule
 \multirow{2}{*}{$Q_2$ } & Security                                            &   12  \\
 & Ethical values and norms                             &   23     \\ \midrule
 \multirow{2}{*}{$Q_3$}  & Political interest and political participation      &   36  \\
 & Political culture and political regimes            &   25   \\ \midrule
\multirow{1}{*}{$C_1$}  & Andorra, Argentina, Australia, Bangladesh, Armenia, Bolivia, Brazil, Myanmar, Canada, Chile, China, Taiwan ROC, Cyprus, Ecuador, Germany, Guatemala, Hong Kong SAR, Indonesia, Iran, Iraq, Japan, Kazakhstan, Jordan, South Korea, Kyrgyzstan, Lebanon, Macao SAR, Maldives, Mongolia, Netherlands, Nicaragua, Pakistan, Philippines, Romania, Russia, Serbia, Singapore, Slovakia, Vietnam, Tajikistan, Turkey, Ukraine, Great Britain, United States, Uruguay, Venezuela & {46} \\ \midrule
\multirow{1}{*}{$C_2$} & Egypt, Ethiopia, Kenya, Libya, Morocco, Nigeria, Tunisia, Zimbabwe & {8} \\ \midrule
\multirow{1}{*}{$C_3$} & Malaysia, Thailand, Czechia, Greece, Nigeria, Morocco, Peru, Colombia, Mexico, Puerto Rico, New Zealand & {11} \\
\bottomrule
\end{tabular}}
\caption{\label{tb:survey_dimensions} Cultural dimensions and question ids of WVS. Question 82-223 and 94-106 are excluded in $Q_1$ as they are demo graphical questions. Demo graphical questions are excluded as they are related to individual attribute regarding participants and does not have relevance to group culture.}
\end{table}

\section{Hyperparameter Settings}\label{app:hyperparams}
Training is performed on a single A100 GPU with a batch size of 16 for Llama3 and Vicuna1.5-7B, and 4 for Vicuna1.5-13B. Both models use the AdamW optimizer with a learning rate of 1e-4 and implement Fully Sharded Data Parallel along with a mixed precision strategy to enhance computational efficiency. In our experiments, we employ LoRA with a rank of 8, a scaling factor lora\_alpha set to 32, and a dropout rate of 0.05.


\paragraph{Model Download} We list all used models here: 

\begin{itemize}
    \item \textbf{Vicuna1.5-7B}: \url{https://huggingface.co/lmsys/vicuna-7b-v1.5}
    \item \textbf{Vicuna1.5-13B}: \url{https://huggingface.co/lmsys/vicuna-13b-v1.5}
    \item \textbf{Distil-Qwen-7B}: \url{https://huggingface.co/deepseek-ai/DeepSeek-R1-Distill-Qwen-7B}
    \item \textbf{Distil-Qwen-14B}: \url{https://huggingface.co/deepseek-ai/DeepSeek-R1-Distill-Qwen-14B}
    \item \textbf{Distil-Qwen-32B}: \url{https://huggingface.co/deepseek-ai/DeepSeek-R1-Distill-Qwen-32B}
    \item \textbf{Llama3-8B-Base}: \url{https://huggingface.co/meta-llama/Meta-Llama-3-8B}
    \item \textbf{Llama3-8B-Instruct}: \url{https://huggingface.co/meta-llama/Meta-Llama-3-8B-Instruct}
    \item \textbf{Qwen-7B}: \url{https://huggingface.co/Qwen/Qwen-7B}
\end{itemize}

\begin{figure}[htbp]
  \centering
  \includegraphics[width=1.0\linewidth]{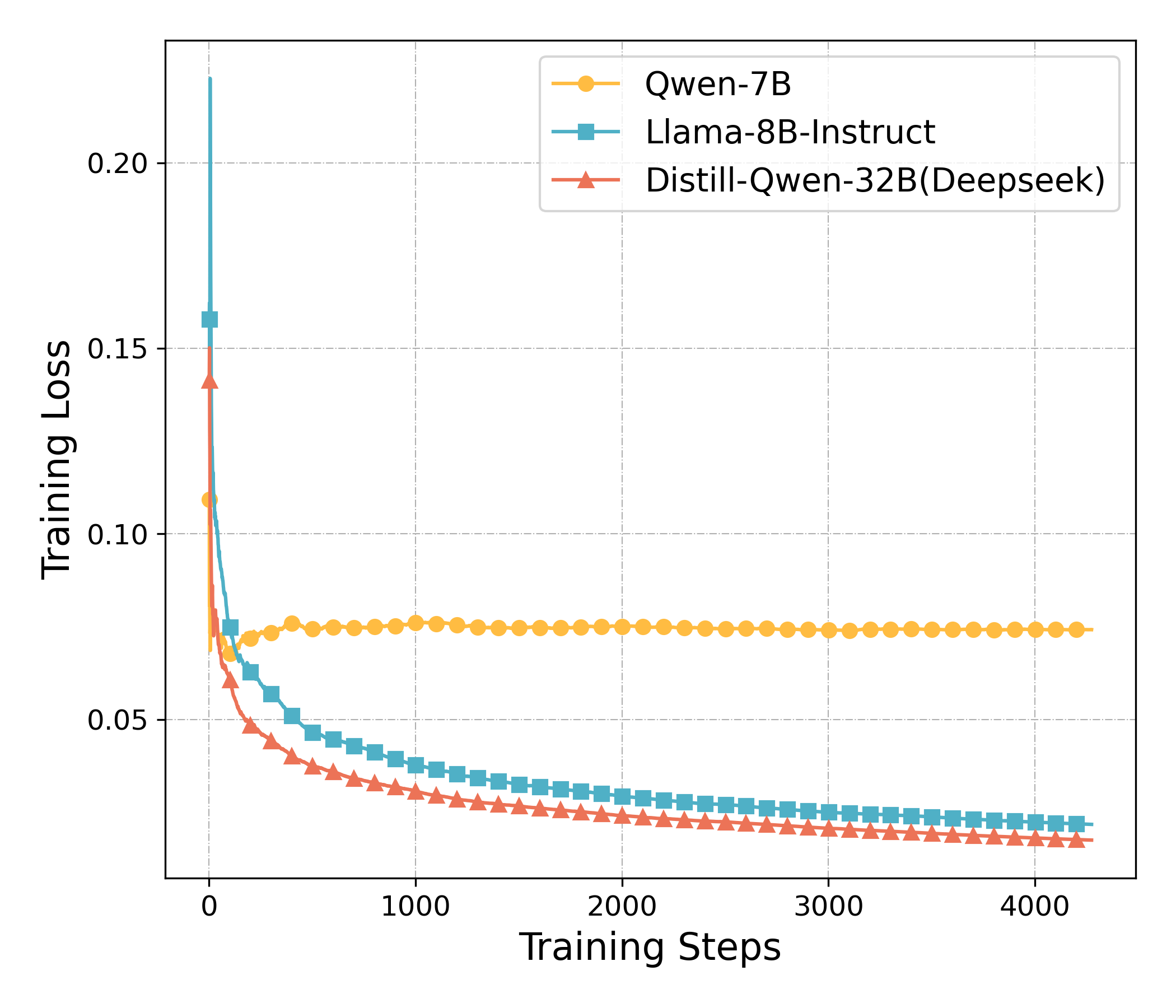} 
  \caption{Training loss comparsion of three models: Qwen-7B, Llama-8B-Instruct, Distill-Qwen-32B(Deepseek)}
  \label{fig:training_loss_comparison}
\end{figure}

\begin{figure*}[t]
    \centering
    \includegraphics[width=0.9\textwidth]{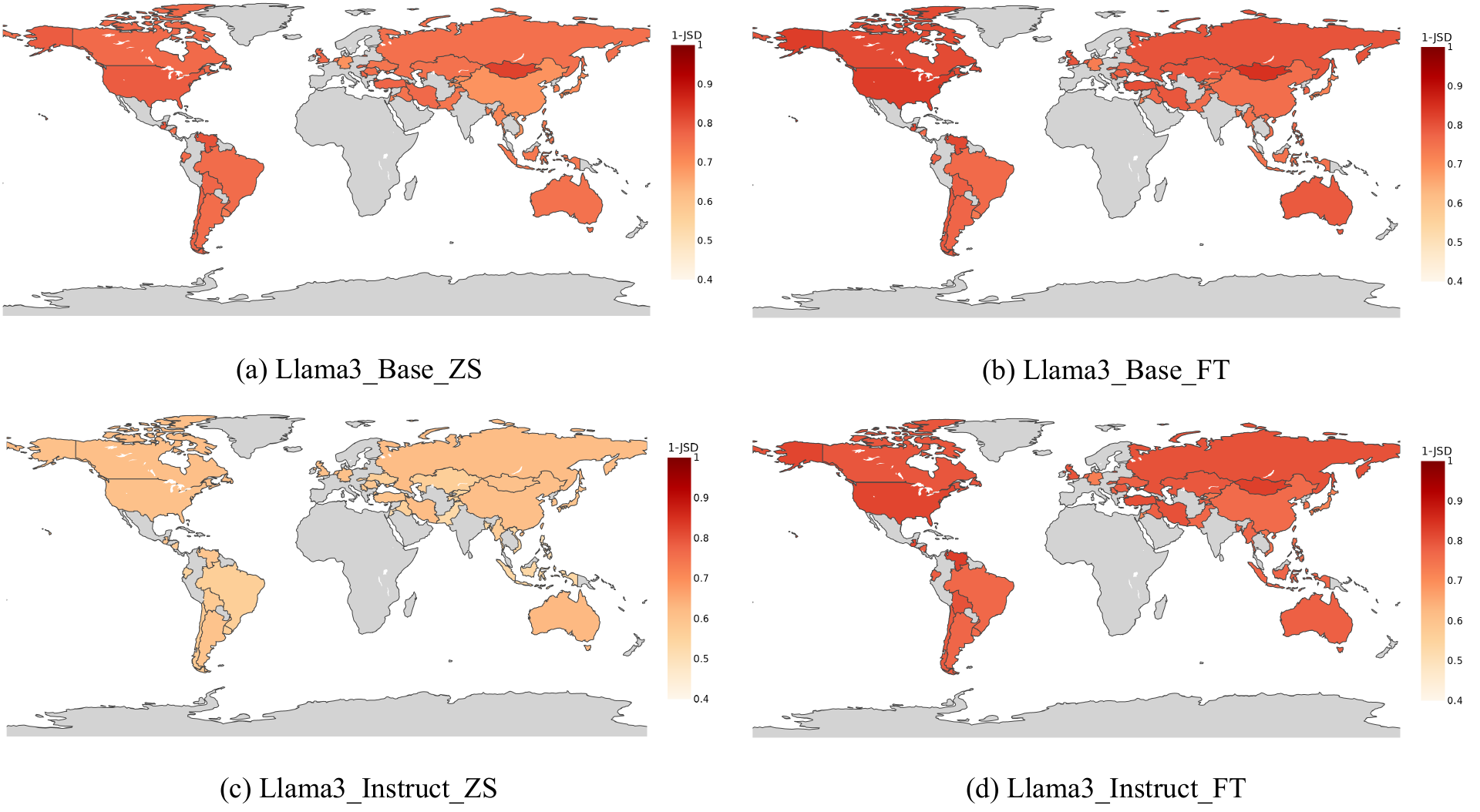} 
    \caption{Distribution of $1-$JSD Global Scores for the Model on Unseen Cultural Questions ($C_1$-$Q_3$). The Instruct model exhibits a more distinct improvement compared to the Base model.}
    \label{fig:global_distrition}
\end{figure*}

\section{More Option Prediction Analysis}

\paragraph{Option Prediction Accuracy.}
In line with Figure~\ref{fig:diversity_distribution}e-\ref{fig:diversity_distribution}f, we present the remaining two sets of Llama3 option prediction accuracy for C3-Q1 and C3-Q3 here (see Figure \ref{fig:acc_wvs_c3}). 

The results indicate that the model performs significantly better on seen questions (C3-Q1) compared to unseen ones (C3-Q3), with Llama3 Instruct models consistently outperforming Base models, and fine-tuned approaches demonstrating superior accuracy over zero-shot methods across most countries. Specifically, the model achieves its highest accuracy in Greece, Nigeria, and Peru, while showing weaker performance in Thailand and Mexico. Despite the challenges in predicting unseen questions, fine-tuning through simulation leads to noticeable improvements, highlighting the potential of fine-tuned models to enhance generalization in novel scenarios.

\paragraph{Country Distributions.} 
Figure~\ref{fig:global_distrition} visualizes the distribution of $1-$JSD scores for the Llama3-Base and Llama3-Instruct models across unseen cultural questions ($C_1$-$Q_3$) in both zero-shot (ZS) and fine-tuned (FT) modes. In ZS mode, both models perform poorly across many regions, with lower scores indicating limited cultural sensitivity. This suggests that without additional tuning, both models struggle to effectively generalize to simulate culturally diverse distributions. 
Furthermore, the Instruct model performs slightly worse, likely due to the reinforcement learning and safety strategies, which may inadvertently reduce its sensitivity to cultures in this scenario.

However, after FT, both models show improvement across all countries, particularly significant in regions previously displaying poor performance in the ZS setting. Additionally, the performance of both the fine-tuned Base model and the Instruct model is observed to be very similar across different countries, demonstrating that our methods can effectively align the Base and Instruct models.

\section{More Baseline Comparsion}
\label{ax:more_baseline}

To further assess the effectiveness of our proposed approach, we compare it against several other baseline methods as follows:

\begin{itemize}
    \item \textbf{KNN}: This method identifies the most similar training question and country (top-1, $k=1$) using BERT embeddings and returns the corresponding option distributions as predictions.
    \item \textbf{Avg\_Culture}: We computes the mean option distribution across all training countries for each known question and adopts a uniform random distribution for unknown questions.
    \item \textbf{JSON-ZS}: This approach prompts the models to directly generate option distributions in JSON format without additional fine-tuning.
\end{itemize}

Table \ref{tab:ex_baseline_results} presents the 1-JSD scores for various baseline methods and our proposed approach. The results indicate that all baseline methods perform substantially worse than our fine-tuned (FT) models. Among the baselines, the JSON-ZS method demonstrates superior performance compared to KNN and Avg\_Culture; however, it remains less effective than both zero-shot (ZS) and fine-tuned (FT) approaches. Notably, fine-tuning consistently yields the highest scores across all evaluated settings, underscoring its effectiveness in improving model performance. 

\begin{table}[h]
    \centering
    \resizebox{\linewidth}{!}{
    \begin{tabular}{lccccccc}
        \toprule
        \textbf{Methods} & \textbf{C1-Q3} & \textbf{C2-Q1} & \textbf{C2-Q3} & \textbf{C3-Q1} & \textbf{C3-Q3} & \textbf{Avg.} \\
        \midrule \midrule
        \textbf{KNN} & 0.381 & 0.518 & 0.371 & 0.541 & 0.384 & 0.439 \\
        \textbf{Avg\_Culture} & 0.360 & 0.509 & 0.348 & 0.518 & 0.368 & 0.421 \\
        \midrule
        \multicolumn{7}{c}{\textit{Llama3-8B-Base}} \\
        \textbf{JSON-ZS} & 0.754 & 0.728 & 0.581 & 0.729 & 0.751 & 0.709 \\
        \textbf{ZS} & 0.749 & 0.768 & 0.759 & 0.781 & 0.770 & 0.765 \\
        \rowcolor{customgray}
        \textbf{FT} & \textbf{0.770} & \textbf{0.858} & \textbf{0.773} & \textbf{0.877} & \textbf{0.781} & \textbf{0.812} \\
        \midrule
        \multicolumn{7}{c}{\textit{Llama3-8B-Instruct}} \\
        \textbf{JSON-ZS} & 0.735 & 0.728 & 0.747 & 0.729 & 0.751 & 0.738 \\
        \textbf{ZS} & 0.585 & 0.650 & 0.589 & 0.657 & 0.584 & 0.613 \\
        \rowcolor{customgray}
        \textbf{FT} & \textbf{0.777} & \textbf{0.881} & \textbf{0.783} & \textbf{0.890} & \textbf{0.784} & \textbf{0.823} \\
        \bottomrule
    \end{tabular}}
    \caption{Comparison of 1-JSD scores across different baseline methods and our approach. Higher values indicate better alignment with the ground-truth distribution.}
    \label{tab:ex_baseline_results}
\end{table}

\begin{figure}[htbp]
  \centering
  \includegraphics[width=0.9\linewidth]{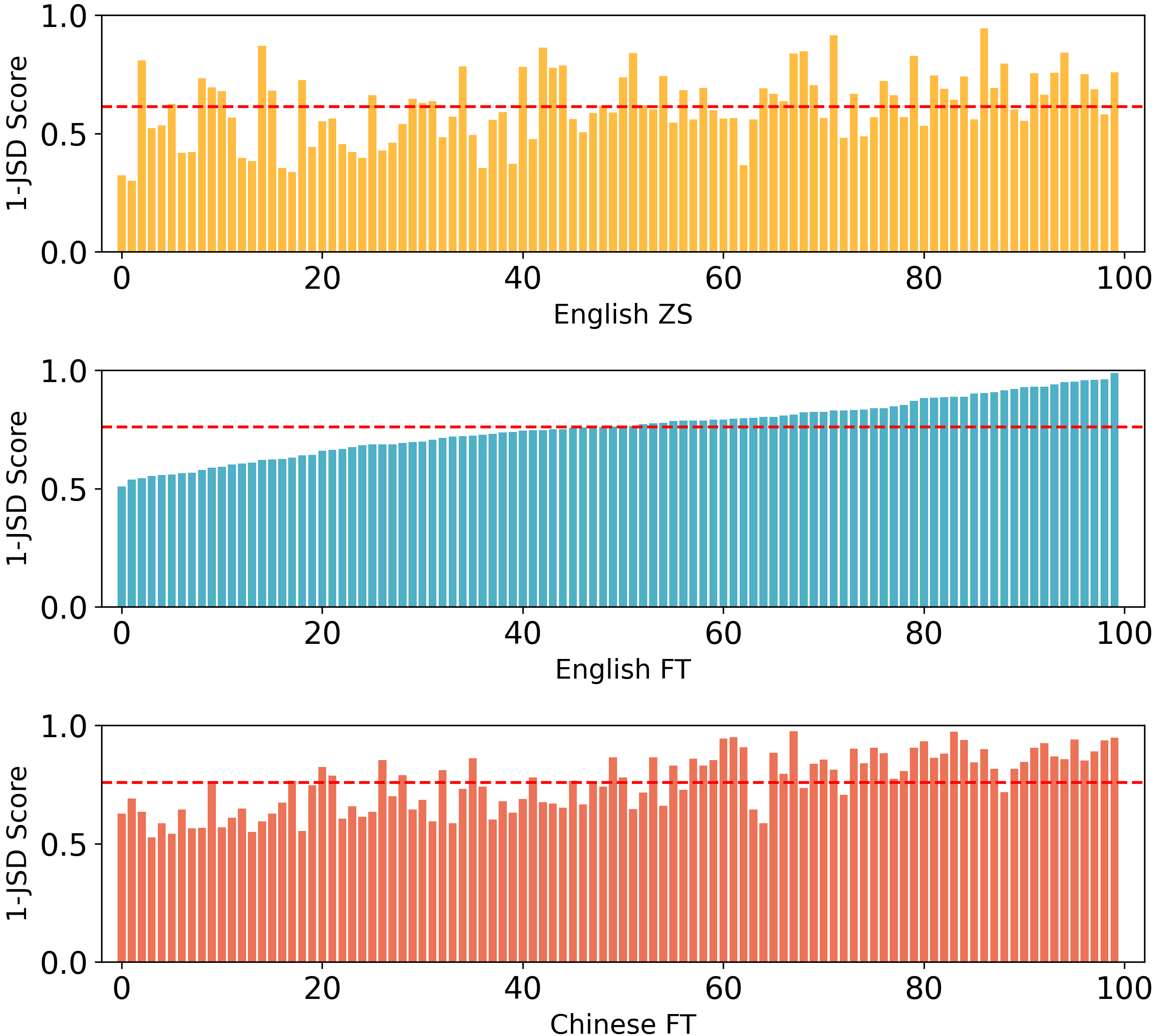} 
  \caption{Comparsion of different languages in 1-JSD score on random selected 100 samples.}
  \label{fig:language_effect}
\end{figure}

\section{More Model Evaluation}
\label{ax:vicuna_results}
In this section, we present additional experiment results on original Qwen model. We observe that the original Qwen models present challenges in training for our task. As shown in Figure \ref{fig:training_loss_comparison}, we compare the training progress of three models. The loss of the Qwen-7B model exhibits difficulty in converging compared to Llama-8B-Instruct and Distill-Qwen-32B. This difficulty may stem from the extensive safety alignment or policy alignment incorporated into Qwen, which could introduce additional constraints or optimization challenges during training.

Secondly, we visualized the 1-JSD score distribution of the model for English ZS, FT, and Chinese FT, as shown in Figure \ref{fig:language_effect}, with the red dashed line indicating the average value. The results show that language differences have a smaller impact on the model than training levels. Additionally, we calculate the Pearson correlation coefficient between English ZS and FT which (0.349) is lower than that between English FT and Chinese FT (0.579). While both correlations are positive, the stronger correlation observed between English FT and Chinese FT suggests that training level exerts a greater influence than language differences.


\end{document}